\documentclass[10pt,twocolumn,letterpaper]{article}

\usepackage{cvpr}
\usepackage{times}
\usepackage{epsfig}
\usepackage{graphicx}
\usepackage{amsmath}
\usepackage{amssymb}
\usepackage{float}
\usepackage{pifont}
\newcommand{\cmark}{\ding{51}}%
\newcommand{\xmark}{\ding{55}}%
\usepackage{xcolor}
\usepackage{booktabs}
\usepackage[percent]{overpic}
\usepackage{authblk}

\makeatletter 
\renewcommand\AB@affilsepx{,   \protect\Affilfont} 
\makeatother

\usepackage[breaklinks=true,bookmarks=false, colorlinks=True, linkcolor=green]{hyperref}

\cvprfinalcopy 

\def\cvprPaperID{6748} 

\begin{document}
\pagenumbering{gobble}

\title{Unsupervised Human Pose Estimation through Transforming Shape Templates}
\author[1]{Luca Schmidtke}
\author[1]{Athanasios Vlontzos}
\author[1]{Simon Ellershaw}
\author[3]{Anna Lukens}
\author[2]{\\Tomoki Arichi}
\author[1]{and Bernhard Kainz}
\affil[1]{Imperial College London}
\affil[2]{King's College London}
\affil[3]{Evelina London Children's Hospital}

\maketitle

\begin{abstract}
Human pose estimation is a major computer vision problem with applications ranging from augmented reality and video capture to surveillance and movement tracking. In the medical context, the latter may be an important biomarker for neurological impairments in infants.
Whilst many methods exist, their application has been limited by the need for well annotated large datasets and the inability to generalize to humans of different shapes and body compositions, \emph{e.g.} children and infants. In this paper we present a novel method for learning pose estimators for human adults and infants in an unsupervised fashion. We approach this as a learnable template matching problem facilitated by deep feature extractors. Human-interpretable landmarks are estimated by transforming a template consisting of predefined body parts that are characterized by 2D Gaussian distributions. Enforcing a connectivity prior guides our model to meaningful human shape representations. We demonstrate the effectiveness of our approach on two different datasets including adults and infants. Project page: \url{infantmotion.github.io}
\end{abstract}

\section{Introduction}
In today's digitized world, images and videos are an almost endless source of unlabeled, but inherently structured data.
Tapping into this reserve of information and knowledge requires the ability to reason in an unsupervised capacity; one of the most compelling and fundamental open problems in machine learning and computer vision. 

Self-supervision approaches have shown evidence that they can provide a good supervisory signal for video data~\cite{jakab_skeleton}. In video recordings an object usually maintains its intrinsic feature distribution but changes its predominantly linear relationships between localized features~\cite{jakab_unsupervised}.

\begin{figure}[t]
\begin{center}
  \includegraphics[width=1.0\linewidth]{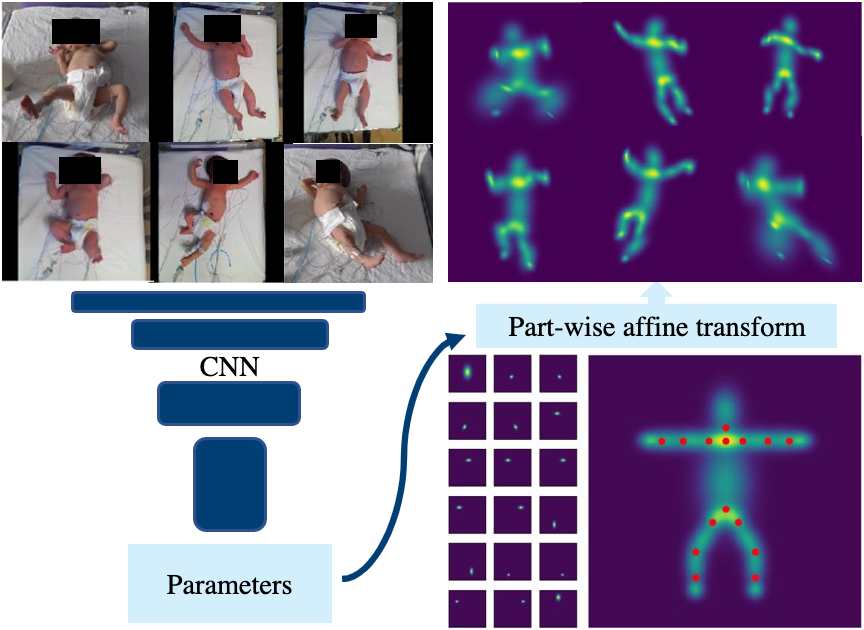}
  \noindent\makebox[\linewidth]{\rule{\linewidth}{0.4pt}}\\
  \vspace{4pt}
  \includegraphics[width=1.0\linewidth]{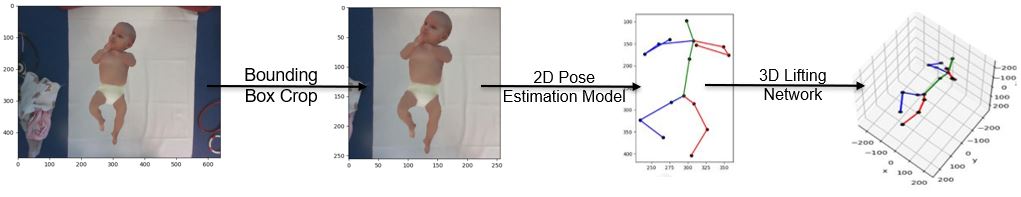}
\end{center}
  \caption{Schematic overview of our approach. Top: We define a part-based human template consisting of 2D Gaussian ellipses and estimate the transformation parameters to estimate the pose of humans with any body composition. Anchor-points are defined between adjacent body parts in order to enforce a connectivity constraint. Bottom: We also evaluate downstream applicability by estimating 3D body poses.}
\label{fig:long}
\label{fig:onecol}
\end{figure}

In this paper we consider the problem of human pose estimation. Motivated by a wide range of applications including motion capture, visual surveillance and robot control, a continuous effort has been put into generating datasets and models where a manually annotated ground truth pose and key points are available as labels for full supervision. These are currently the most attractive approaches for industrial applications due to their promise of higher accuracy. 

However, ground truth generation is laborious and often limited  to a narrow domain, for example, standard poses of healthy adults. In domains with limited demand or special requirements, extensive labeling efforts are often not justified. Such domains include medical applications, where key point definitions may vary according to diagnostic aims and body shapes might not comply with the learned expectations from a standard training set. Indeed, motion tracking has a variety of applications in medicine, for example to examine the progression of neurological disease and to evaluate treatment success~\cite{kontschieder2014quantifying}, the assessment of injury~\cite{bae2009level} or for the early diagnosis of impaired neurological development in infants~\cite{autism, palsy}. None of these applications allow for excessive data collection and annotation, often because of a limited number of subjects, restrictions on recording in the clinical environment and economical considerations. Moreover, the direct application of models trained on common benchmark datasets~\cite{h36m,mpii_dataset,he2016deep} is often challenging. 

To tackle this issue, unsupervised and weakly supervised pose estimation methods~\cite{jakab_unsupervised, lorenz, Kundu} decompose images into \textit{appearance}, which encodes individual differences such as clothing or body height and \textit{pose}, describing the individual's positioning and configuration of limbs and joints with a canonical latent code. In this context, self-supervision tasks, such as image reconstruction or translation, have been shown to be powerful tools to  
estimate pose as a \textit{factor of variation} across images instead of relying on strong, manual supervision signals.

We therefore aim at learning the 2D geometry of object categories such as humans and infants with no additional supervision. We exploit the structured information provided by raw videos of continuous pose changes and propose to control inductive bias directly for arbitrary object categories through the manual definition of very simple templates.
Thus, we intend to automatically train a neural representation that can predict the 2D pose from a single input image. 
We show that if such a 2D pose prediction is accurate and compliant with an expected shape prior, these estimations can be extrapolated to 3D poses with a lower error than other existing methods.

We  present a method for the unsupervised estimation of 2D keypoints requiring only a simple template and an unannotated video of a single human performing actions in front of a static background to learn a meaningful pose representation.
Inspired  by previous work \cite{jakab_skeleton}, where this problem is framed as a \emph{conditional image generation and translation} approach, pose information is utilized to recover a particular frame of a video from  any other randomly chosen time-point. 
Despite the effective use of self-supervision and representational bottleneck, this approach still requires another prior in the form of unpaired labels and introduces susceptibility to domain shift if these labels come from a different datset. Our model however does not require an additional dataset of unpaired 2D pose examples and relies solely on a simple 2D template consisting of connected body parts modeled as 2D Gaussians. The update of these Gaussians can be learned as affine transformations. Even though images are 2D representations of 3D information, affine transformations allow to model all possible projected configurations of body parts. Motivated by part-based approaches such as \cite{puppet} we introduce anchor-points in order to enforce connectivity between body parts and regularize model training and prediction.

In summary, we make the following contributions:

\begin{itemize}
    \item We introduce a conceptually simple but effective method to learn 2D human-interpretable keypoints based on transforming a single manually defined 2D template.
    
    \item Our proposed approach is capable of performing 2D human pose estimation without any  additional need for labeled data, either paired or unpaired.

    \item We demonstrate the high adaptability of our approach by evaluating it on benchmark data and in the wild on a challenging infant pose estimation dataset.
\end{itemize}

\section{Related Work}
We consider the problem of predicting the 2D pose of an object from a single 2D RGB image as a pose recognition task. We structure related methods according to \textbf{full supervision}, where dense manual annotations are paired to each frame in a dataset, and \textbf{weak supervision or no supervision}, when only partial  or no annotations are available or when models are transferred to a new domain. The question when a method can be referred to as weakly supervised or unsupervised remains debatable, which is why we do not make this distinction explicit.

Inductive bias and prior knowledge are also important in this context. Very often latent manifold distributions that are known a priori are used as an initialization step. This can include models that have been learned from other data or any other type of supervision or empirical priors. Similar to recent attempts in this field, \emph{e.g.} \cite{jakab_unsupervised}, our method is unsupervised but it uses a very simple empirical prior, which we hypothesize leads to better results.

\paragraph{Fully supervised} models rely on carefully annotated data, which is available for narrow fields of applications, \emph{e.g.} MS COCO keypoints~\cite{coco_dataset}, MPII Human Pose database~\cite{mpii_dataset}, Human3.6M~\cite{h36m} or LSP~\cite{he2016deep}. Methods utilizing these datasets are usually trained without additional priors due to the abundance of direct labels. Pictorial structures~\cite{felzenszwalb2005pictorial, andriluka2009pictorial, ouyang2014multi, pishchulin2013poselet, ramakrishna2014pose, sapp2010adaptive, yang2011articulated} have been used to describe poses and CNNs have shown evidence to be powerful estimators for keypoints~\cite{toshev2014deeppose} and their uncertainties~\cite{heatmaps}. Confidence heatmaps are popular for scenes where a single pose needs to be estimated~\cite{belagiannis2017recurrent, bulat2016human, carreira2016human, hourglass, pfister2015flowing, tompson2015efficient, wei2016convolutional} or multiple poses at once~\cite{cao2017realtime, insafutdinov2016deepercut}. Our framework does not use existing image annotation to learn a pose prediction model.

\paragraph{Weakly supervised and unsupervised} methods including self-supervised approaches have grown in popularity due to their efficiency in dealing with limited ground truth data. This becomes especially important when there are no large publicly available datasets available for the target domain.
Jakab et al. \cite{jakab_unsupervised} propose to learn a pose representation via conditional image translation. By leveraging the extracted pose representation and given another frame containing the same person in a different pose they task the model to reconstruct the original image. However, a pixel-wise reconstruction loss does not encourage meaningful, low-dimensional representations. To avoid the model from simply encoding pixel information similar to an autoencoder, a representational bottleneck in the form of $k$ tuples of $(x,y)$ coordinates is introduced. This enables unsupervised pose landmark detection provided that the background remains static. However, the extracted landmarks follow no common convention and are difficult or impossible to interpret. To tackle this, \cite{jakab_skeleton} further propose the introduction of a fully differentiable image-based representation, resembling a human skeleton. Adversarial training with unpaired pose labels is required to make the model converge towards human-interpretable outputs.

Zhang et al. \cite{zhang} introduce equivariance and invariance constraints under an autoencoder-based formulation, while Lorenz et al. \cite{lorenz} expand this approach by introducing disentanglement between object pose and appearance. Kundu et. al. proposed to use an energy-based optimization approach combined with a part-based shape template to estimate 3D poses in images with varying backgrounds \cite{Kundu}. The method produces impressive results but still relies on a set of unpaired real 3D pose labels.

Earlier methods learn to predict dense 3D human meshes from sparse 2D keypoint annotations, \emph{e.g.}~\cite{kanazawa2018end}, by using a parametric human mesh
model~\cite{loper2015smpl} and regularization by adversarial learning from motion capture coordinates. Related approaches have been proposed in~\cite{gecer2019ganfit, geng20193d, gerig2018morphable, mehta2017monocular, sengupta2018sfsnet, tung2017adversarial, wang2019adversarial, yang20183d}.

Others propose methods to match pairs of images of an object, but sacrifice geometric invariants such as keypoints to achieve this~\cite{kanazawa2016warpnet,rocco2017convolutional,shu2018deforming}. Sparse and dense landmarks are introduced by~\cite{sundermeyer2018implicit,thewlis2017unsupervised,thewlis2019unsupervised} in the unsupervised context. Synthetic views of 3D models like in \cite{sundermeyer2018implicit,hesse_ferns} are a reasonable workaround which our method does not use. 

Other noteworthy approaches learn a dense deformation field, \emph{e.g.} for faces~\cite{shu2018deforming, wiles2018self}. In contrast to their methods and similar to~\cite{jakab_skeleton} we predict semantically meaningful keypoints, however, our points can be freely defined and easily changed through a simple template. Thus, the quality of our landmarks is higher per definition of our approach and adaptable for any given application. 


\paragraph{Human pose estimation in medicine:}
 Medical applications pose a larger challenge for pose estimation algorithms. 
For infants, existing learned pose estimation models suffer from domain shift, thus, focused models have been developed. Pose Estimation in videos of infants has been proposed as an early diagnosis tool of diseases related to impaired neurological development affecting the sensorimotor system~\cite{autism, palsy}. In previous work, marker-based approaches relying on optical \cite{Airaksinen2020} or electromagnetic \cite{PMID:18707688} tracking have been utilized to track motion in infants. However, these methods rely on clinical specialists, expertise, often costly equipment and a substantial amount of manual preparation and calibration. Different marker-less approaches have been proposed based on optical flow \cite{Stahl2012AnOF} or particle matching \cite{Rahmati2015WeaklySM}. With the advent of compact and cheap cameras with integrated depth sensors, several more recent works combine images and depth information with random fern classifiers \cite{hesse_ferns}, a deformable parts model \cite{Khan} or by employing a shape model \cite{Olsen2014ModelBasedMT,hesse2018learning}.

For medical applications 3D pose prediction holds additional value.
The most common approach to 3D pose estimation is full supervision as detailed above. One of the earliest approaches outlining a supervised deep learning approach to the task was proposed by Li et al. \cite{li20143d} with a network analogous to~\cite{toshev2014deeppose}. 
 In \cite{chen20173d}, the authors proposed splitting the human pose estimation task into two parts. The first is a generic 2D pose estimation model using CNNs, the second is a non-parametric nearest neighbor model that paired the estimated 2D pose to the closest 2D pose from a dictionary of paired 2D and 3D poses. Martinez et al.~\cite{martinez2017simple} took this a step further,  replacing the 3D dictionary lookup model with a deep learning 3D lifting network that took the estimated 2D pose vectors as an input and produced 3D 'lifted' outputs. The use of a differentiable soft argmax function \cite{chapelle2010gradient} allows end-to-end training of a fully differentiable model and an L2 loss function can be used to directly regress 3D keypoint locations.

\begin{figure}[t]
\begin{center}
  \includegraphics[width=1.0\linewidth]{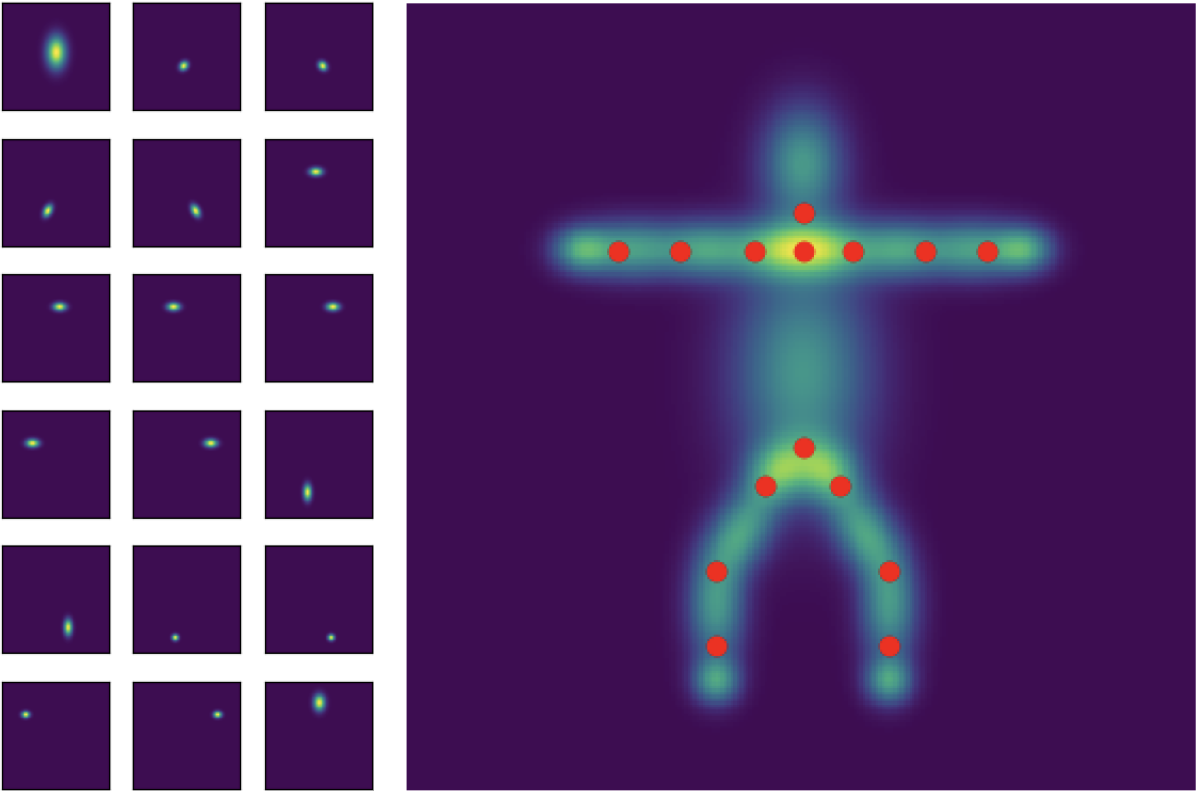}
\end{center}
  \caption{Left: Body part heatmaps; Right: combined template with defined anchor points plotted in red.}
\label{fig:template}
\end{figure}

\section{Approach}
\begin{figure*}[t]
\begin{center}
  \includegraphics[width=1.0\linewidth]{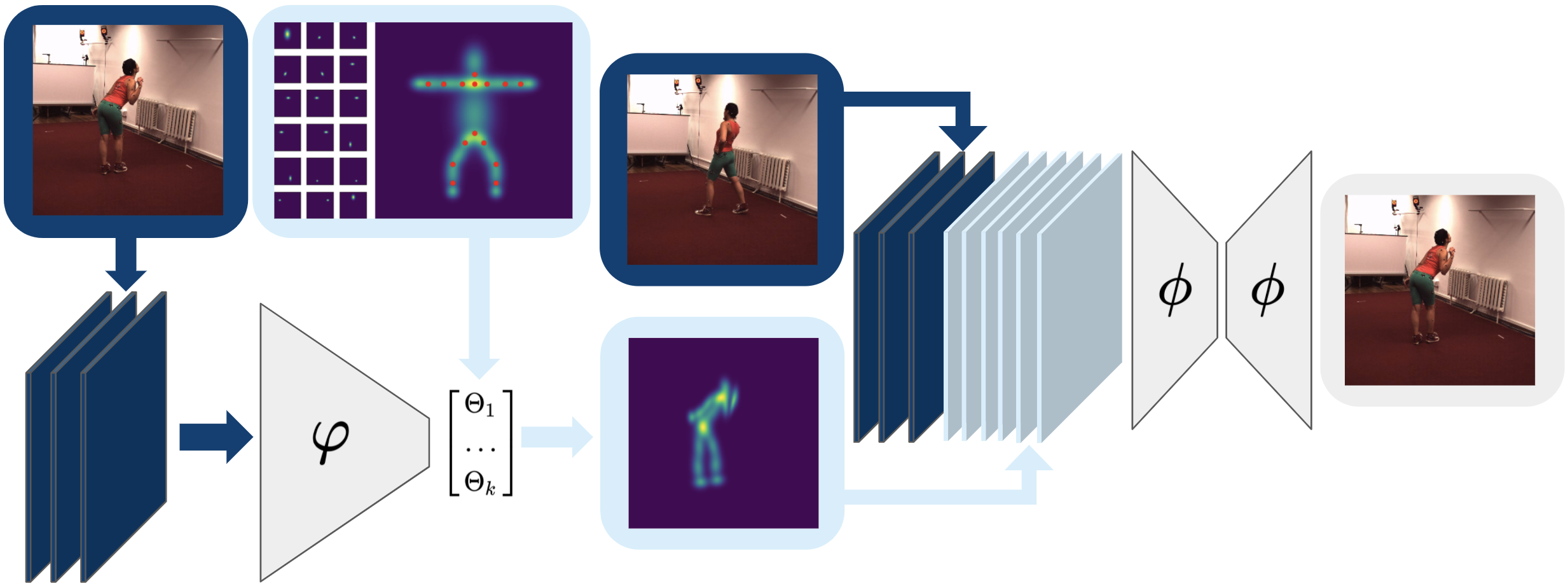}
\end{center}
  \caption{Schematic overview of our approach: We define a part-based human puppet template and predict the transformation parameters to estimate the pose for an input frame.}
\label{fig:method}
\end{figure*}
In this section we present our proposed method. 
We draw inspiration from~\cite{jakab_skeleton} and formulate a self-supervision task by tasking our model to reconstruct an initial frame $\mathbf{f}_t$ conditioned on an estimated low-dimensional pose representation $\Tilde{p}_t$. On a  high level, our method is composed of two modules. The first network $\varphi$, given the initial frame $\mathbf{f}_t$ and  a template $\mathbf{T}$  consisting of 2D Gaussian heatmaps of human body parts,  extracts an estimate $\Tilde{p}_t$ of the true pose $p_t$ represented by a spatial arrangement of different body parts. The second module is an encoder-decoder network $\phi$ which receives as input the previously extracted pose together with a frame $\mathbf{f}_{t+k}$ containing the same person in a different pose $p_{t+k}$, and tasked to reconstruct the original frame $f_t$.
\begin{align}
&\Tilde{\mathbf{f}}_t = \phi(\mathbf{f}_{t+k}, \Tilde{p}_t) \\
&\Tilde{p}_t = \varphi(\mathbf{f}_t, \mathbf{T})
\end{align}
The pose representation $\tilde{p}_t$ is found by optimizing the model to encode all necessary information to recover the original pose whilst enforcing representational constraints to avoid  the encoding of low-level image features.

\subsection{Template and Anchor-points}
Inspired by ~\cite{Kundu} and \cite{puppet}, we design a human template as shown in Figure~\ref{fig:template} that consists of 18 body parts in an effort to represent human anatomy (forearm, torso, head, etc.). Instead of transforming this template according to previously found keypoints in the image, we directly estimate the transformations and make use of the template as a strong prior. Each part is characterized by its central position ($x_0$, $y_0$) surrounded by a 2D Gaussian with variance ($\mathrm{Var}_x$, $\mathrm{Var}_y$). We pre-define the mean and variance of each Gaussian for all body parts in an effort to represent the length and width of the individual parts. Hence, we are able to model a canonical T-pose as shown in Figure~\ref{fig:template}. This pose was chosen because it roughly represents the mean of possible pose configurations.
Each representation of a body part $\mathbf{l} \in \mathbb{R}^{h \times w}$ is saved in an individual channel resulting in a tensor of shape $(B, K, H, W)$ where $B$ is the batch size, $K$ the number of body parts and $H, W$ are width and height of the input image.

For each body part $\mathbf{l}_j$ in our template $\mathbf{T}$ we define $N_j$ (up to three depending on which part) anchor points $\mathbf{a}_i^j \in \mathbb{R}^2$ in image space, with $j=1,..., K$ and $i=1,...,N_j$. Our template design and the resulting anchor points coincide with the most commonly used landmark definitions for human pose estimation. However, the detected landmarks can be easily changed by simply choosing different points on each body part.

\subsection{Pose Extractor\label{pose_extractor}}
The network  $\varphi: \mathbb{R}^{3\times h \times w} \rightarrow \mathbb{R}^{k \times 3 \times 3}$ is implemented as a  fully-convolutional neural network followed by a fully-connected layer. An image $\mathbf{f}_t$ is passed through a series of  down-sampling convolutional layers and mapped to a pose representation described by the parameters of affine transformations $\Theta_k \in \mathbb{R}^{3x3}$ for each $k$-th body part from the original template:
\begin{equation}
\Theta = 
\begin{bmatrix}
\theta_{1,1} & \theta_{1,2} & t_x\\
\theta_{2,1} & \theta_{2,2} & t_y \\
0 & 0 & 1
\end{bmatrix},
\end{equation}
where $t_{x,y}$ represent the translations across the two dimensions and $\theta$ correspond to rotation, shear, and scale. Each part of the template is then warped by its corresponding transform, resulting in $\mathbf{T'} = (\vartheta_0 (\mathbf{l}_0),..., \vartheta_k (\mathbf{l}_k))$, where $\vartheta_i$ is the corresponding transform for each $\Theta_i$.
\subsection{Image Translation Module}
The network $\phi: \mathbb{R}^{3+k \times h \times w} \rightarrow \mathbb{R}^{3 \times h \times w}$ receives a different frame $\mathbf{f}_{t+k}$ along with the transformed template $\mathbf{T'}$ concatenated along the channel dimension and outputs another image $\Tilde{\mathbf{f}}_t$ following an encoder-decoder pathway. Intuitively, this network learns to implicitly disentangle content (\textit{i.e.}, person identity) and pose in an image and transforms the content according to the conditioning pose.

\subsection{Training Objectives}
Our training objective is split into three parts: A reconstruction loss is associated with the image translation module while a boundary and an anchor loss guide the training of the pose extractor network. 

\paragraph{Reconstruction Loss}
The main loss component in our method is the reconstruction objective between the original frame $\mathbf{f}_t$ and the output of the image translation network $\phi$. Similar to \cite{jakab_skeleton}, we found it to be beneficial to use a perceptual loss with a pretrained VGG network in order to stabilize training:
\begin{equation}
    \mathcal{L}_{recon} = \vert\vert \psi(\Tilde{\mathbf{f}}_t) - \psi(\mathbf{f_t}) \vert \vert_1^1,
\end{equation}
where ${\psi(\mathbf{f}})$ are feature vectors extracted from a frame by the VGG network and $\Tilde{\mathbf{f}}_t$ the output of the image translation network.

\paragraph{Anchor-point Loss}
Each anchor point is being transformed by the corresponding body part transform: $\Tilde{\mathbf{a}}_i^j = \Theta_j \mathbf{a}_i^j$.
In order to enforce connectivity between body parts, we define a set 
$A=\{(\tilde{\mathbf{a}}_j^l, \tilde{\mathbf{a}}_k^m)\vert l \ne m \}$
containing all pairs of  transformed anchor points which we require to have the same position. Lastly we compute the mean squared L2 distance of the two points over all tuples:
\begin{equation}
    \mathcal{L}_{\text{anchor}} = \dfrac{1}{M}\sum_{\tilde{\mathbf{a}}_j^l, \tilde{\mathbf{a}}_k^m  \in A} \vert\vert \tilde{\mathbf{a}}_j^l -\tilde{\mathbf{a}}_k^m\vert\vert_2^2,
\end{equation}
where $M=$ number of anchor point pairs.

\paragraph{Boundary Loss}
Additionally, we found it beneficial for convergence and to prevent the network from outputting transformation parameters leading to a translation of the template shapes out of the image boundaries to enforce the anchor points to be contained within the image:
\begin{equation}
    \mathcal{L}_{bx}=
    \begin{cases}
      \vert a_{i, x}^j \vert, & \text{if}\ \vert a_{i, x}^j\vert > B \\
      0, & \text{otherwise},
    \end{cases}
\end{equation}

\noindent where $a_{i, x}^j$ is the x-coordinate of the anchor point  in pixel space and B corresponds to the size of the (quadratic) image. The loss $\mathcal{L}_{by}$ for the y-component is analogous.

\begin{figure*}[t]
\begin{center}
  \includegraphics[width=1.0\linewidth]{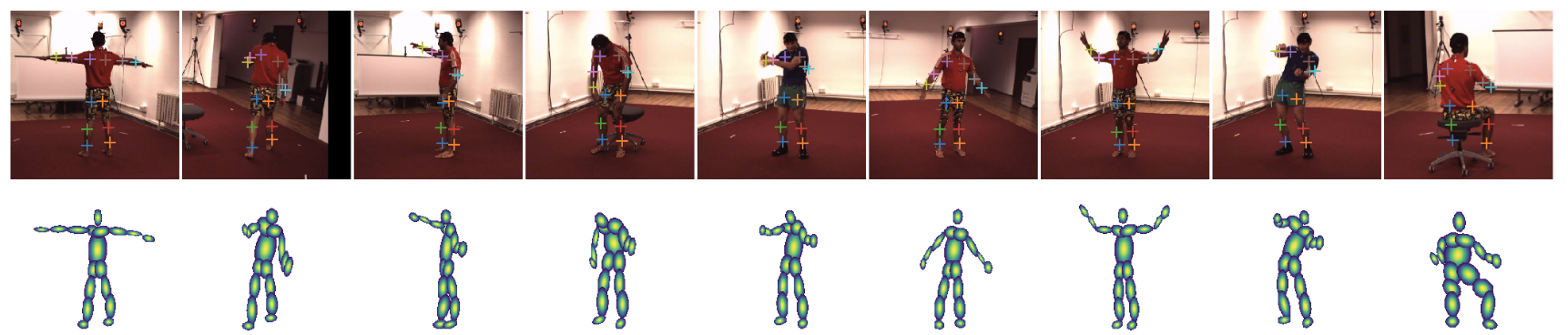}
\end{center}
  \caption{Results of our model on random frames from subjects 9 and 11 in Human 3.6m. Top row: input images with detected keypoints. Bottom row: corresponding deformed shape templates}
\label{fig:results1}
\end{figure*}

\paragraph{Combined Loss Formulation}
In summary our training objective is expressed as 
\begin{equation}
    \mathcal{L} = \mathcal{L}_{\text{recon}} + \lambda_1 \: \mathcal{L}_{\text{anchor}} + \lambda_2 \: (\mathcal{L}_{bx} + \mathcal{L}_{by}).
\end{equation}

\paragraph{Training}
Our whole model including all subnetworks is trained in an end-to-tend manner. For sampling pairs, we define a course grid and corresponding bounding boxes based on the supplied masks and select frames from the same bounding box to ensure a static background. Note, that the hyper-parameters $\lambda_1,\lambda_2$ are empirically tuned.

\begin{figure}[]
\begin{center}
  \includegraphics[width=1.0\linewidth]{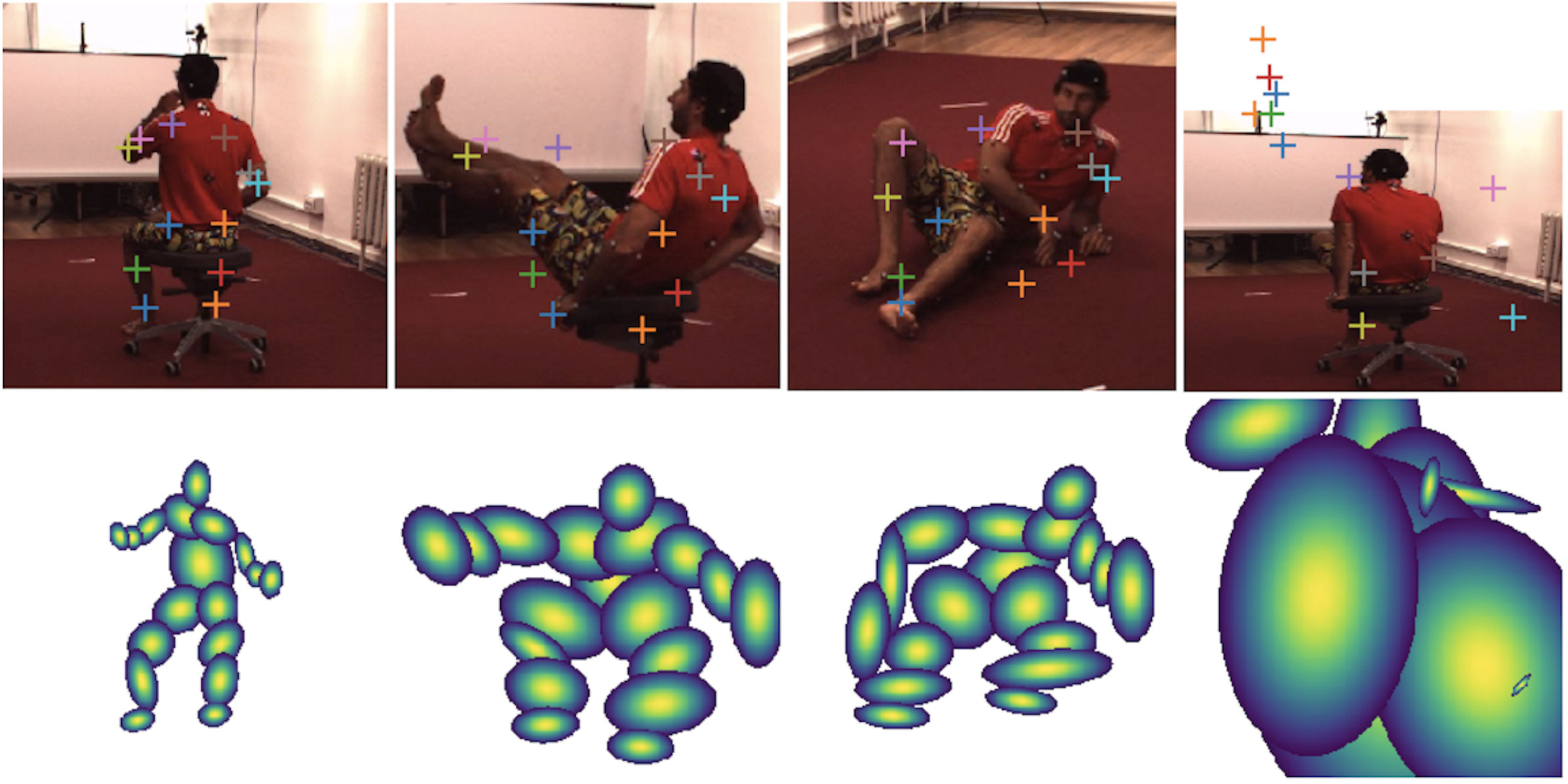}
\end{center}
  \caption{Example for limitations of our model. While sitting is  a very difficult pose, facing away from the camera is the most challenging situation. We observe the full loss of coherent landmarks in some of these cases. Moreover, the model switches left and right when a person is facing away from the camera.}
\label{fig:failures}
\end{figure}

\section{Experiments}
We evaluate our approach on two different datasets, Human3.6m and an infant dataset from clinical practice. We compare our method to results with various degrees of supervision, as reported in recent literature. Additionally, we performed an ablation study to demonstrate the effectiveness of our proposed anchor and boundary losses.

\paragraph{Human3.6m} is an industry standard dataset for human pose estimation \cite{h36m} containing 3.6 million images with corresponding 3D and 2D landmarks and captured with different actors in a controlled studio environment with a static background.

\paragraph{Infant Dataset} is a clinical dataset containing videos of 24 infants captured in their cod in the hospital. The dataset was recorded and curated by the authors and their clinical collaborators using an image labelling tool \cite{labelbox}. All subjects were recruited from consenting parents while the study and its use has been cleared by the appropriate ethics committees. We train our method on 20 subjects and evaluate on four, resulting in a total of 290k unlabelled training and 471 manually labelled test images.

\paragraph{Moving Infants In RGB-D (MINI-RGBD)} consists of 12 synthetically rendered movement sequences of infants with different body shapes and backgrounds \cite{hesse2018computer}. The dataset was designed to cover challenging and diverse movements and overall contains 12k images together with various label information including landmarks and masks. We introduce this dataset to demonstrate the issue of domain shift when using these labels as an additional prior.

\begin{figure*}[t]
\begin{center}
  \includegraphics[width=1.0\linewidth]{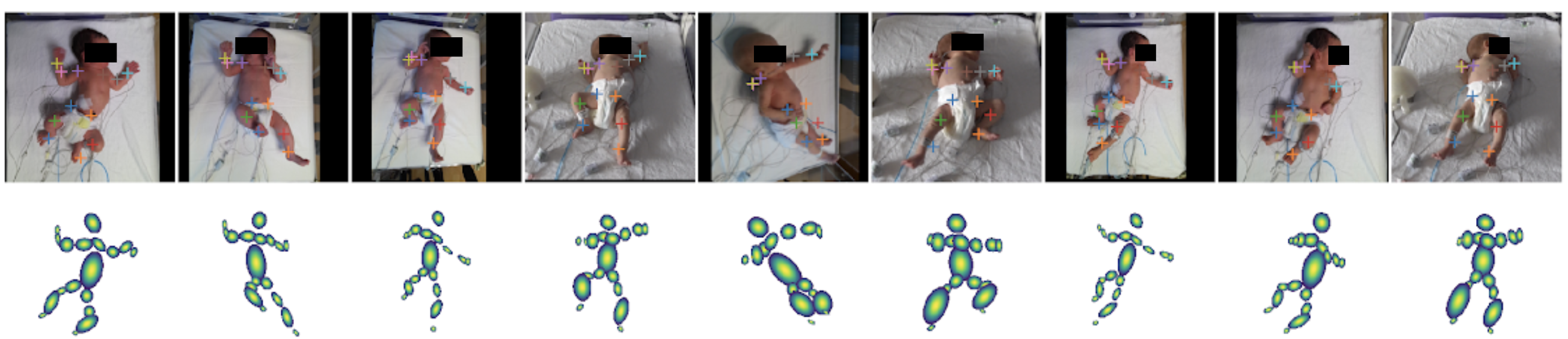}
\end{center}
  \caption{Results of our model on random frames from the test subjects of our in-house dataset. Anonymisation was applied to maintain patient privacy. Top row: input images with detected keypoints. Bottom row: corresponding deformed shape templates}
\label{fig:results_infants}
\end{figure*}

\subsection{Evaluation Procedure}
\paragraph{Human 3.6m}
In order to make our results comparable to recent literature with regards to the Human3.6m dataset, we adopt the evaluation strategy from \cite{lorenz, zhang}. The model is trained on subjects 1,5,6,7,8 and tested on subject 9. We restrict the performed actions to \{direction, discussion, posing, waiting, greeting, walking\} during testing in order to ensure mostly upright poses, resulting in a total of 80k images for testing. 16 landmarks are predicted and compared with the ground truth. More results can be found in the supplemental material. The comparison methods make use of the supplied estimated rough person masks and substract the background. We did  not perform this extra preprocessing step.

Many proposed unsupervised or weakly supervised methods require an additional post-processing step and direct supervision via ground truth 2D keypoints. Most use a linear regressor, in order to output human-interpretable landmarks. Our approach returns interpretable landmarks in form of anchorpoints \textit{by design}. The latter also coincide with the most commonly used conventions, since they are placed around the joints.

\paragraph{Infant Dataset}
As a baseline, we fine-tune a fully supervised ResNet based network \cite{pose_baselines} pretrained on both ImageNet and adult poses from MPII Human Pose \cite{mpii_dataset}. For further comparison, we implemented and trained the method proposed by Jakab et al.~\cite{jakab_skeleton} on our clinical infant images combined with unpaired poses from the synthetic MINI-RGBD infant dataset \cite{hesse2018computer} in order to demonstrate the method's susceptibility to domain shift. 12 landmarks are predicted and compared with the ground truth.

\begin{table}[]
\centering

\begin{tabular}{l|cccc}
       & BBox & SPP & UL & T \\ \hline
Lorenz \cite{lorenz} & \textcolor{red}{\cmark}    & \textcolor{red}{\cmark}   & \textcolor{green}{\xmark}  & \textcolor{green}{\xmark} \\
Zhang \cite{zhang}  & \textcolor{red}{\cmark}   & \textcolor{red}{\cmark}   & \textcolor{green}{\xmark}  & \textcolor{green}{\xmark} \\
Jakab \cite{jakab_skeleton}  & \textcolor{red}{\cmark}   & \textcolor{green}{\xmark}  & \textcolor{red}{\cmark}  & \textcolor{green}{\xmark} \\
Ours   & \textcolor{red}{\cmark}  & \textcolor{green}{\xmark}  & \textcolor{green}{\xmark}  & \textcolor{red}{\cmark}
\end{tabular}
\caption{\label{table:1} Comparing different sources of labels used for different methods. \textbf{BBox}: bounding box centered around the person, \textbf{SPP}: supervised post-processing. A linear regressor is applied to map discovered landmarks to human-interpretable locations. \textbf{UL}: Use of unpaired manually annotated poses as a prior. \textbf{T}: the manual design of a single template including individual body parts and anchorpoints}
\end{table}

For  a fair comparison of the results, we make this explicit by adding \textit{supervised post-processing} in parenthesis for all results where this step was included. We also compiled Table~\ref{table:1} to clearly indicate the different sources of labels that are used by the different methods.

\setlength\tabcolsep{3pt}
\begin{table}[]
\centering
\begin{tabular}{lccccccc}
\toprule
H36M    & all  & wait & pose & greet & direct & discuss & walk \\ \midrule
\multicolumn{8}{c}{\textbf{\textit{fully supervised baseline}}} \\
Newell \cite{hourglass} & 2.16 & 1.88 & 1.92 & 2.15  & 1.62   & 1.88    & 2.21 \\
\midrule
\multicolumn{8}{c}{\textbf{\textit{self-supervised + supervised post-processing}}} \\
Thewlis \cite{thewlis2017unsupervised} & 7.51 & 7.54 & 8.56 & 7.26  & 6.47   & 7.93    & 5.40 \\
Zhang \cite{zhang}   & 4.14 & 5.01 & 4.61 & 4.76  & 4.45   & 4.91    & 4.61 \\
Lorenz \cite{lorenz}  & 2.79 & --    & --    & --     & --      & --       & --    \\
\midrule
\multicolumn{8}{c}{\textbf{\textit{self-supervised (unpaired labels)}}} \\
Jakab \cite{jakab_skeleton}   & 2.73 & 2.66 & 2.27 & 2.73  & 2.35   & 2.35    & 4.00 \\
\midrule
\multicolumn{8}{c}{\textbf{\textit{self-supervised (template, no labels)}}} \\
Ours    & 3.31 & 3.51 & 3.28 & 3.50  & 3.03   & 2.97   & 3.55 \\ 
\bottomrule
\end{tabular}
\caption{\label{table:results} Comparison with state-of-the-art methods for human landmark detection on the Simplified Human3.6M dataset; \%-MSE normalized by image size is reported on a per action basis. Note that our method does not require any annotations at all while results are en-par with the state-of-the-art unsupervised approaches utilizing unpaired labels or post-processing.}
\end{table}

\setlength\tabcolsep{1.5pt}
\begin{table}[]
\centering
\begin{tabular}{lccccccc}
\toprule
Infants & all  & hips & knees & feet & shoulders & hands & params\\ \midrule
\multicolumn{8}{c}{\textbf{\textit{fully supervised (fine-tuned) baseline}}} \\
Xiao \cite{pose_baselines}    & 1.74 & 2.39 & 1.50  & 1.47 & 1.76      & 1.59 &34.0 M  \\
\midrule
\multicolumn{8}{c}{\textbf{\textit{self-supervised (unpaired labels)}}} \\
Jakab \cite{jakab_skeleton}   & 8.98 & 6.89 & 8.18  & 13.15 & 5.33      & 11.36 &8.6 M  \\
\midrule
\multicolumn{8}{c}{\textbf{\textit{self-supervised (template, no labels)}}} \\
Ours    & 4.86 & 3.79 & 4.60  & 5.53 & 3.19      & 7.21 & 7.8 M  \\ \bottomrule
\end{tabular}
\caption{\label{table:results_inf} Comparison with state-of-the-art methods for the Infant dataset; \%-MSE normalized by image size is reported on a per body-part basis. Our method outperforms prior unsupervised approaches for this task because it is not influenced by annotation domain shift.}
\label{table:infant_results}
\end{table}

\subsection{Results}
\begin{figure}[b]
\begin{center}
  \includegraphics[width=1\linewidth]{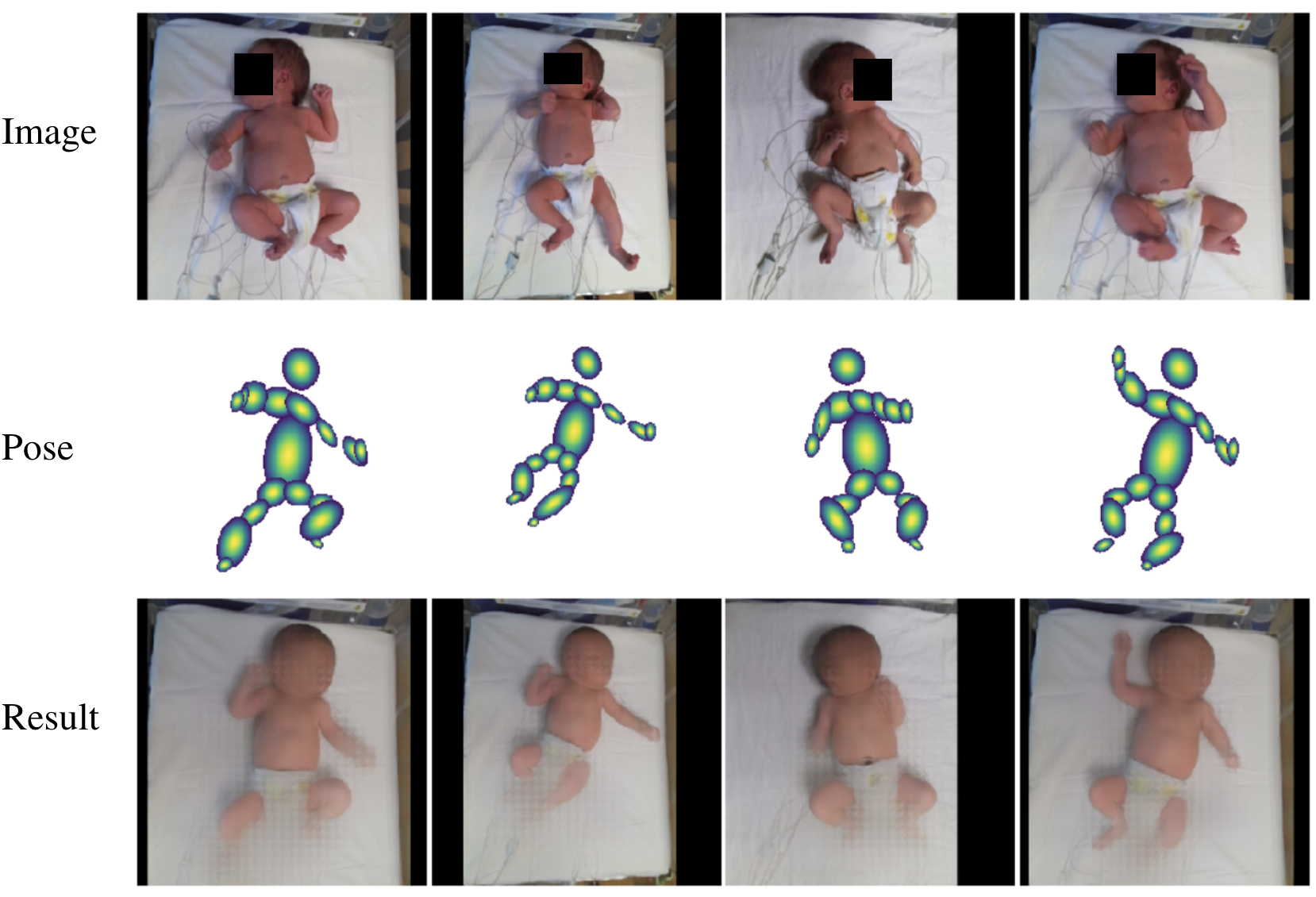}
\end{center}
  \caption{Pose conditioning. Top row: Input image; Middle row: New Conditioning Pose; Bottom row: Resulting image. We observe that our image translation sub-module is able to disentangle pose and content and produce photo-realistic images with the conditioning poses.}
\label{fig:translation}
\end{figure}
\paragraph{Human 3.6m}
For adult pose estimation, we summarize our results in Table~\ref{table:results}. As expected, none of the self-supervised methods perform as well as the supervised baseline due to the lack of labels. All prior work makes use of paired or unpaired manual annotations in some capacity. Despite the complete lack of such annotations, our method performs competitively on the same task while only requiring the template. Our model is also able to predict landmarks for difficult poses such as sitting on a chair, as can be seen in Figure~\ref{fig:results1}. Errors are largest when self-occlusion occurs. In the most severe cases, when a person is facing away from the camera and both arms and hands are covered by the body, estimating these landmarks becomes extremely difficult without further supervision in the form of manually annotated examples or additional views captured by a second camera.



\paragraph{Infant Pose Estimation}
The results for infant pose estimation are summarized in Table~\ref{table:infant_results} and \ref{table:3d_infant_results_}.  Figure~\ref{fig:results_infants} displays predictions of our model on infants. Our implemented version of \cite{jakab_skeleton} performs worse despite the access to 11,000 unpaired landmark annotations. We attribute this to domain shift. Since the model relies on adversarial training on these labels, the performance will drop if the latter are not covering a diverse enough range of possible poses. In fact, the authors demonstrate a drop in performance in their own experiments when using labels from a different dataset.

Our model is capable of predicting consistent and interpretable landmarks from images of infants with different body shapes and in different poses. Again, the largest errors are introduced by self-occlusion, especially when arms are positioned in front of the chest and shoulders. This is consistent with our observation that the landmark detection works best on the legs, which are most of the time not positioned above or below other body parts.
\\
\noindent \textbf{Ablation study}
In order to verify the individual contribution of our loss components, we partially train several models with different loss configurations and present the results in Table \ref{table:results_abl}.
\begin{table}[h]
    \centering
    \begin{tabular}{cccccc}
    \toprule
     all  & hips & knees & feet & shoulders & hands\\ \midrule
    \multicolumn{6}{c}{\textbf{\textit{anchor and boundary loss (ours)}}} \\
     6.54 & 3.56 & 5.19  & 8.90 & 4.70      & 7.95 \\
    \midrule
    \multicolumn{6}{c}{\textbf{\textit{with anchor, no boundary loss}}} \\
     65.99 & 64.65 & 71.70  & 76.33 & 59.00     & 62.59 \\
    \midrule
    \multicolumn{6}{c}{\textbf{\textit{with boundary, no anchor loss}}} \\
     12.42 &13.84 & 13.17 & 7.29 & 12.94      & 8.76 \\
    \midrule
    \multicolumn{6}{c}{\textbf{\textit{no anchor, no boundary loss (model diverges)}}} \\
     446.47 & 433.64 & 385.77  & 662.41 & 502.69      & 276.49 \\
    \bottomrule
    \end{tabular}
    \caption{\label{table:results_abl}Mean joint distance in \% of image size (values$>$100 are outside of image). All models trained for 10 epochs.}
\end{table}

\noindent \textbf{Image Translation}
Since our decoder network is trained to synthesize images conditioned on the pose input, we are able to perform image editing by combining a frame $\mathbf{f}$ with a different set of poses $p$. The images displayed in Figure~\ref{fig:translation} illustrate the principle and demonstrate disentanglement between pose and appearance. Moreover, the image translation modules' ability to produce images given a conditioning pose could serve as a tool in the analysis for indicators of impaired neurological development. This, however, remains a topic of future work.  

\paragraph{Limitations} Our method is limited by occlusions and partial views of the body and is currently not able to distinguish back-facing from front-facing poses. If not accounted for, this can result in large errors due to the switching of joints with their left/right counterpart. The problem only occurs for adult pose estimation, as infants are lying on their backs during a recording and are physically not able to turn around. We believe that the inclusion of a directional vector could alleviate the observed limitation. 

\section{Downstream application study}

\begin{figure}[]
\begin{center}
\begin{overpic}[width=1\linewidth]{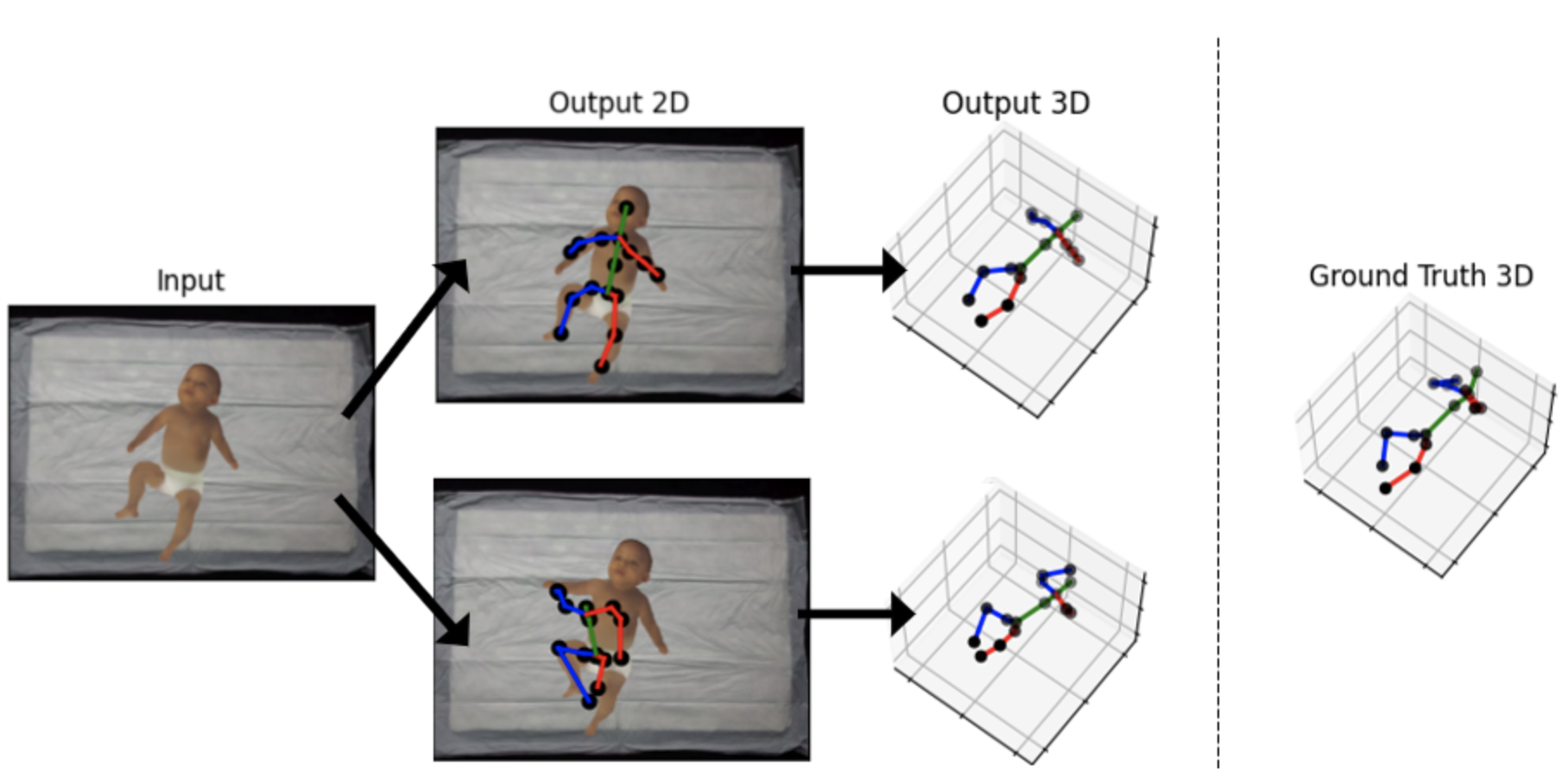} 
\put (17,39) {\textbf{\textit{Ours}}}
\put (-2,1) {\textbf{\textit{Jakab et al.~\cite{jakab_skeleton}}}}
\end{overpic}
 \end{center}
  \caption{Visual results from the lifting network on a synthetic infant image. Top: ours, bottom: Jakab et al.~\cite{jakab_skeleton}. The accuracy of the predicted 2D keypoints has a noticeable  impact on the final 3D predictions.}
\label{fig:app1}
\end{figure}

To study the downstream effects of 2D pose prediction accuracy in a realistic setting we integrate our method into a setup as it is currently used for clinical evaluation of neonatal movement quality. Physiological movement quality assessment requires 3D coordinates, thus a fully supervised 3D lifting network is applied after 2D pose estimation. The lifting model is adapted from the work of Martinez et al.~\cite{martinez2017simple}.    
This model was first pre-trained on the MPI\_INF\_3DHP adult dataset \cite{mehta2017monocular} before fine-tuning on the infant MINI-RGBD dataset~\cite{hesse2018computer}. Figure~\ref{fig:app1} compares the the qualitative differences when results from previous 2D pose estimation methods~\cite{jakab_skeleton} are used or ours. Quantitatively this can be captured by evaluating the joint position error in mm. 
These results are summarized in Table~\ref{table:3d_infant_results_}.

\begin{table}[t]
\centering
\begin{tabular}{@{}lcccccc@{}}
\toprule
Synthetic Infants & all & \multicolumn{1}{l}{hips} & \multicolumn{1}{l}{knees} & \multicolumn{1}{l}{feet} & \multicolumn{1}{l}{shoulders} & \multicolumn{1}{l}{hands} \\ \midrule
\multicolumn{7}{c}{\textbf{\textit{self-supervised (unpaired labels)}}}               \\
Jakab~\cite{jakab_skeleton} & \multicolumn{1}{c}{59.7} & 10.1 & 58.4 & 73.3 & 58.0 & 101.5 \\ \midrule
\multicolumn{7}{c}{\textbf{\textit{self-supervised (template, no labels)}}}           \\
Ours  &                     44.7 & 8.0 & 57.5 & 80.9 & 35.6 & 78.7 \\ \bottomrule
\end{tabular}
  \caption{\label{table:3d_infant_results_}3D lifting results. The error is reported as the distance between predicted and ground truth 3D landmarks in mm.}
\end{table}

\section{Conclusion}
We presented a novel approach to unsupervised human pose estimation by transforming a part-based shape template. Given a video of a person moving in front of a static background, we are able to predict human interpretable keypoints without requiring any paired or unpaired labels. We have exhibited our method's performance on two different datasets and achieved  similar or better results in unsupervised human pose estimation while only requiring a simple canonical 2D pose template instead of numerous manual labels. Moreover, our method is able to adapt to humans or infants of different shapes and sizes and alternative keypoints can be defined easily by choosing different locations on each body part. We also demonstrate an increase in performance when using our method's extracted poses in downstream tasks like 3D pose extrapolation, which is of significance for both industrial and medical applications. Although some manual work is required to define the template, this could be easily realized in a simple GUI, where the user would be able to draw body parts and define anchor- as well as keypoints within minutes. 

\noindent\textbf{Acknowledgements:}
 supported by EPSRC EP/S013687/1.

{\small
\bibliographystyle{ieee_fullname}
\bibliography{egbib}

\begin{thebibliography}{10}\itemsep=-1pt

\bibitem{labelbox}
Labelbox.
\newblock {\em Online, Available: https://labelbox.com}, 2021.

\bibitem{Airaksinen2020}
Manu Airaksinen, Okko R{\"a}s{\"a}nen, Elina Il{\'e}n, Taru H{\"a}yrinen, Anna
  Kivi, Viviana Marchi, Anastasia Gallen, Sonja Blom, Anni Varhe, Nico
  Kaartinen, Leena Haataja, and Sampsa Vanhatalo.
\newblock Automatic posture and movement tracking of infants with wearable
  movement sensors.
\newblock {\em Scientific Reports}, 2020.

\bibitem{mpii_dataset}
Mykhaylo Andriluka, Leonid Pishchulin, Peter Gehler, and Bernt Schiele.
\newblock 2d human pose estimation: New benchmark and state of the art
  analysis.
\newblock In {\em IEEE Conference on Computer Vision and Pattern Recognition
  (CVPR)}, June 2014.

\bibitem{andriluka2009pictorial}
Mykhaylo Andriluka, Stefan Roth, and Bernt Schiele.
\newblock Pictorial structures revisited: People detection and articulated pose
  estimation.
\newblock In {\em 2009 IEEE conference on computer vision and pattern
  recognition}, pages 1014--1021. IEEE, 2009.

\bibitem{bae2009level}
TS Bae, K Choi, and M Mun.
\newblock Level walking and stair climbing gait in above-knee amputees.
\newblock {\em Journal of medical engineering \& technology}, 33(2):130--135,
  2009.

\bibitem{belagiannis2017recurrent}
Vasileios Belagiannis and Andrew Zisserman.
\newblock Recurrent human pose estimation.
\newblock In {\em 2017 12th IEEE International Conference on Automatic Face \&
  Gesture Recognition (FG 2017)}, pages 468--475. IEEE, 2017.

\bibitem{bulat2016human}
Adrian Bulat and Georgios Tzimiropoulos.
\newblock Human pose estimation via convolutional part heatmap regression.
\newblock In {\em European Conference on Computer Vision}, pages 717--732.
  Springer, 2016.

\bibitem{cao2017realtime}
Zhe Cao, Tomas Simon, Shih-En Wei, and Yaser Sheikh.
\newblock Realtime multi-person 2d pose estimation using part affinity fields.
\newblock In {\em Proceedings of the IEEE conference on computer vision and
  pattern recognition}, pages 7291--7299, 2017.

\bibitem{carreira2016human}
Joao Carreira, Pulkit Agrawal, Katerina Fragkiadaki, and Jitendra Malik.
\newblock Human pose estimation with iterative error feedback.
\newblock In {\em Proceedings of the IEEE conference on computer vision and
  pattern recognition}, pages 4733--4742, 2016.

\bibitem{chapelle2010gradient}
Olivier Chapelle and Mingrui Wu.
\newblock Gradient descent optimization of smoothed information retrieval
  metrics.
\newblock {\em Information retrieval}, 13(3):216--235, 2010.

\bibitem{chen20173d}
Ching-Hang Chen and Deva Ramanan.
\newblock 3d human pose estimation= 2d pose estimation+ matching.
\newblock In {\em Proceedings of the IEEE Conference on Computer Vision and
  Pattern Recognition}, pages 7035--7043, 2017.

\bibitem{felzenszwalb2005pictorial}
Pedro~F Felzenszwalb and Daniel~P Huttenlocher.
\newblock Pictorial structures for object recognition.
\newblock {\em International journal of computer vision}, 61(1):55--79, 2005.

\bibitem{palsy}
Fabrizio Ferrari, Giovanni Cioni, Christa Einspieler, Maria~Federica Roversi,
  Arend~F Bos, Paola~Bruna Paolicelli, Andrea Ranzi, and Heinz R.~F. Prechtl.
\newblock Cramped synchronized general movements in preterm infants as an early
  marker for cerebral palsy.
\newblock {\em Archives of pediatrics \& adolescent medicine}, 156 5:460--7,
  2002.

\bibitem{gecer2019ganfit}
Baris Gecer, Stylianos Ploumpis, Irene Kotsia, and Stefanos Zafeiriou.
\newblock Ganfit: Generative adversarial network fitting for high fidelity 3d
  face reconstruction.
\newblock In {\em Proceedings of the IEEE Conference on Computer Vision and
  Pattern Recognition}, pages 1155--1164, 2019.

\bibitem{geng20193d}
Zhenglin Geng, Chen Cao, and Sergey Tulyakov.
\newblock 3d guided fine-grained face manipulation.
\newblock In {\em Proceedings of the IEEE Conference on Computer Vision and
  Pattern Recognition}, pages 9821--9830, 2019.

\bibitem{gerig2018morphable}
Thomas Gerig, Andreas Morel-Forster, Clemens Blumer, Bernhard Egger, Marcel
  Luthi, Sandro Sch{\"o}nborn, and Thomas Vetter.
\newblock Morphable face models-an open framework.
\newblock In {\em 2018 13th IEEE International Conference on Automatic Face \&
  Gesture Recognition (FG 2018)}, pages 75--82. IEEE, 2018.

\bibitem{he2016deep}
Kaiming He, Xiangyu Zhang, Shaoqing Ren, and Jian Sun.
\newblock Deep residual learning for image recognition.
\newblock In {\em Proceedings of the IEEE conference on computer vision and
  pattern recognition}, pages 770--778, 2016.

\bibitem{hesse2018computer}
Nikolas Hesse, Christoph Bodensteiner, Michael Arens, Ulrich~G Hofmann, Raphael
  Weinberger, and A~Sebastian Schroeder.
\newblock Computer vision for medical infant motion analysis: State of the art
  and rgb-d data set.
\newblock In {\em European Conference on Computer Vision}, pages 32--49.
  Springer, 2018.

\bibitem{hesse2018learning}
Nikolas Hesse, Sergi Pujades, Javier Romero, Michael~J. Black, Christoph
  Bodensteiner, Michael Arens, Ulrich~G. Hofmann, Uta Tacke, Mijna
  Hadders-Algra, Raphael Weinberger, Wolfgang M\"uller-Felber, and A.~Sebastian
  Schroeder.
\newblock Learning an infant body model from {RGB-D} data for accurate full
  body motion analysis.
\newblock 2018.

\bibitem{hesse_ferns}
N. {Hesse}, G. {Stachowiak}, T. {Breuer}, and M. {Arens}.
\newblock Estimating body pose of infants in depth images using random ferns.
\newblock {\em 2015 IEEE International Conference on Computer Vision Workshop
  (ICCVW)}, 2015.

\bibitem{insafutdinov2016deepercut}
Eldar Insafutdinov, Leonid Pishchulin, Bjoern Andres, Mykhaylo Andriluka, and
  Bernt Schiele.
\newblock Deepercut: A deeper, stronger, and faster multi-person pose
  estimation model.
\newblock In {\em European Conference on Computer Vision}, pages 34--50.
  Springer, 2016.

\bibitem{h36m}
Catalin Ionescu, Dragos Papava, Vlad Olaru, and Cristian Sminchisescu.
\newblock Human3.6m: Large scale datasets and predictive methods for 3d human
  sensing in natural environments.
\newblock {\em IEEE Transactions on Pattern Analysis and Machine Intelligence},
  36(7):1325--1339, jul 2014.

\bibitem{jakab_unsupervised}
Tomas Jakab, Ankush Gupta, Hakan Bilen, and Andrea Vedaldi.
\newblock Unsupervised learning of object landmarks through conditional image
  generation.
\newblock In S. Bengio, H. Wallach, H. Larochelle, K. Grauman, N. Cesa-Bianchi,
  and R. Garnett, editors, {\em Advances in Neural Information Processing
  Systems 31}, pages 4016--4027. Curran Associates, Inc., 2018.

\bibitem{jakab_skeleton}
Tomas Jakab, A. Gupta, Hakan Bilen, and A. Vedaldi.
\newblock Self-supervised learning of interpretable keypoints from unlabelled
  videos.
\newblock {\em 2020 IEEE/CVF Conference on Computer Vision and Pattern
  Recognition (CVPR)}, pages 8784--8794, 2020.

\bibitem{kanazawa2018end}
Angjoo Kanazawa, Michael~J Black, David~W Jacobs, and Jitendra Malik.
\newblock End-to-end recovery of human shape and pose.
\newblock In {\em Proceedings of the IEEE Conference on Computer Vision and
  Pattern Recognition}, pages 7122--7131, 2018.

\bibitem{kanazawa2016warpnet}
Angjoo Kanazawa, David~W Jacobs, and Manmohan Chandraker.
\newblock Warpnet: Weakly supervised matching for single-view reconstruction.
\newblock In {\em Proceedings of the IEEE Conference on Computer Vision and
  Pattern Recognition}, pages 3253--3261, 2016.

\bibitem{PMID:18707688}
Dominik Karch, Keun-Sun Kim, Katarzyna Wochner, Joachim Pietz, Hartmut
  Dickhaus, and Heike Philippi.
\newblock Quantification of the segmental kinematics of spontaneous infant
  movements.
\newblock {\em Journal of biomechanics}, 41(13), September 2008.

\bibitem{Khan}
Muhammad~Hassan Khan, Manuel Schneider, Muhammad~Shahid Farid, and Marcin
  Grzegorzek.
\newblock Detection of infantile movement disorders in video data using
  deformable part-based model.
\newblock {\em Sensors}, 18, 2018.

\bibitem{kontschieder2014quantifying}
Peter Kontschieder, Jonas~F Dorn, Cecily Morrison, Robert Corish, Darko Zikic,
  Abigail Sellen, Marcus D’Souza, Christian~P Kamm, Jessica Burggraaff,
  Prejaas Tewarie, et~al.
\newblock Quantifying progression of multiple sclerosis via classification of
  depth videos.
\newblock In {\em International Conference on Medical Image Computing and
  Computer-Assisted Intervention}, pages 429--437. Springer, 2014.

\bibitem{Kundu}
Jogendra~Nath Kundu, S. Seth, V. Jampani, M. Rakesh, R. Babu, and Anirban
  Chakraborty.
\newblock Self-supervised 3d human pose estimation via part guided novel image
  synthesis.
\newblock {\em 2020 IEEE/CVF Conference on Computer Vision and Pattern
  Recognition (CVPR)}, pages 6151--6161, 2020.

\bibitem{li20143d}
Sijin Li and Antoni~B Chan.
\newblock 3d human pose estimation from monocular images with deep
  convolutional neural network.
\newblock In {\em Asian Conference on Computer Vision}, pages 332--347.
  Springer, 2014.

\bibitem{coco_dataset}
Tsung-Yi Lin, Michael Maire, Serge Belongie, James Hays, Pietro Perona, Deva
  Ramanan, Piotr Doll{\'a}r, editor="Fleet~David Zitnick, C.~Lawrence", Tomas
  Pajdla, Bernt Schiele, and Tinne Tuytelaars.
\newblock Microsoft coco: Common objects in context.
\newblock In {\em Computer Vision -- ECCV 2014}, pages 740--755. Springer
  International Publishing, 2014.

\bibitem{loper2015smpl}
Matthew Loper, Naureen Mahmood, Javier Romero, Gerard Pons-Moll, and Michael~J
  Black.
\newblock Smpl: A skinned multi-person linear model.
\newblock {\em ACM transactions on graphics (TOG)}, 34(6):1--16, 2015.

\bibitem{lorenz}
Dominik Lorenz, Leonard Bereska, Timo Milbich, and B. Ommer.
\newblock Unsupervised part-based disentangling of object shape and appearance.
\newblock {\em 2019 IEEE/CVF Conference on Computer Vision and Pattern
  Recognition (CVPR)}, pages 10947--10956, 2019.

\bibitem{martinez2017simple}
Julieta Martinez, Rayat Hossain, Javier Romero, and James~J Little.
\newblock A simple yet effective baseline for 3d human pose estimation.
\newblock In {\em Proceedings of the IEEE International Conference on Computer
  Vision}, pages 2640--2649, 2017.

\bibitem{mehta2017monocular}
Dushyant Mehta, Helge Rhodin, Dan Casas, Pascal Fua, Oleksandr Sotnychenko,
  Weipeng Xu, and Christian Theobalt.
\newblock Monocular 3d human pose estimation in the wild using improved cnn
  supervision.
\newblock In {\em 2017 International Conference on 3D Vision (3DV)}, pages
  506--516. IEEE, 2017.

\bibitem{hourglass}
Alejandro Newell, Kaiyu Yang, and Jia Deng.
\newblock Stacked hourglass networks for human pose estimation.
\newblock In Bastian Leibe, Jiri Matas, Nicu Sebe, and Max Welling, editors,
  {\em Computer Vision -- ECCV 2016}, pages 483--499, Cham, 2016. Springer
  International Publishing.

\bibitem{Olsen2014ModelBasedMT}
Mikkel~Damgaard Olsen, Anna Herskind, Jens~Bo Nielsen, and Rasmus~R. Paulsen.
\newblock Model-based motion tracking of infants.
\newblock 2014.

\bibitem{ouyang2014multi}
Wanli Ouyang, Xiao Chu, and Xiaogang Wang.
\newblock Multi-source deep learning for human pose estimation.
\newblock In {\em Proceedings of the IEEE Conference on Computer Vision and
  Pattern Recognition}, pages 2329--2336, 2014.

\bibitem{pfister2015flowing}
Tomas Pfister, James Charles, and Andrew Zisserman.
\newblock Flowing convnets for human pose estimation in videos.
\newblock In {\em Proceedings of the IEEE International Conference on Computer
  Vision}, pages 1913--1921, 2015.

\bibitem{pishchulin2013poselet}
Leonid Pishchulin, Mykhaylo Andriluka, Peter Gehler, and Bernt Schiele.
\newblock Poselet conditioned pictorial structures.
\newblock In {\em Proceedings of the IEEE Conference on Computer Vision and
  Pattern Recognition}, pages 588--595, 2013.

\bibitem{Rahmati2015WeaklySM}
Hodjat Rahmati, Ralf Dragon, Ole~Morten Aamo, Lars Adde, {\O}yvind Stavdahl,
  and Luc~Van Gool.
\newblock Weakly supervised motion segmentation with particle matching.
\newblock {\em Computer Vision and Image Understanding}, 2015.

\bibitem{ramakrishna2014pose}
Varun Ramakrishna, Daniel Munoz, Martial Hebert, James~Andrew Bagnell, and
  Yaser Sheikh.
\newblock Pose machines: Articulated pose estimation via inference machines.
\newblock In {\em European Conference on Computer Vision}, pages 33--47.
  Springer, 2014.

\bibitem{rocco2017convolutional}
Ignacio Rocco, Relja Arandjelovic, and Josef Sivic.
\newblock Convolutional neural network architecture for geometric matching.
\newblock In {\em Proceedings of the IEEE conference on computer vision and
  pattern recognition}, pages 6148--6157, 2017.

\bibitem{sapp2010adaptive}
Benjamin Sapp, Chris Jordan, and Ben Taskar.
\newblock Adaptive pose priors for pictorial structures.
\newblock In {\em 2010 IEEE Computer Society Conference on Computer Vision and
  Pattern Recognition}, pages 422--429. IEEE, 2010.

\bibitem{sengupta2018sfsnet}
Soumyadip Sengupta, Angjoo Kanazawa, Carlos~D Castillo, and David~W Jacobs.
\newblock Sfsnet: Learning shape, reflectance and illuminance of facesin the
  wild'.
\newblock In {\em Proceedings of the IEEE Conference on Computer Vision and
  Pattern Recognition}, pages 6296--6305, 2018.

\bibitem{shu2018deforming}
Zhixin Shu, Mihir Sahasrabudhe, Riza Alp~Guler, Dimitris Samaras, Nikos
  Paragios, and Iasonas Kokkinos.
\newblock Deforming autoencoders: Unsupervised disentangling of shape and
  appearance.
\newblock In {\em Proceedings of the European conference on computer vision
  (ECCV)}, pages 650--665, 2018.

\bibitem{Stahl2012AnOF}
Ayelet Stahl, Christian Schellewald, Oyvind Stavdahl, Ole~Morten Aamo, Lars
  Adde, and Harald Kirker{\o}d.
\newblock An optical flow-based method to predict infantile cerebral palsy.
\newblock {\em IEEE Transactions on Neural Systems and Rehabilitation
  Engineering}, 20:605--614, 2012.

\bibitem{sundermeyer2018implicit}
Martin Sundermeyer, Zoltan-Csaba Marton, Maximilian Durner, Manuel Brucker, and
  Rudolph Triebel.
\newblock Implicit 3d orientation learning for 6d object detection from rgb
  images.
\newblock In {\em Proceedings of the European Conference on Computer Vision
  (ECCV)}, pages 699--715, 2018.

\bibitem{autism}
Philip Teitelbaum, Osnat Teitelbaum, J Nye, Joshua~B. Fryman, and Ralph~G.
  Maurer.
\newblock Movement analysis in infancy may be useful for early diagnosis of
  autism.
\newblock {\em Proceedings of the National Academy of Sciences of the United
  States of America}, 95 23:13982--7, 1998.

\bibitem{thewlis2019unsupervised}
James Thewlis, Samuel Albanie, Hakan Bilen, and Andrea Vedaldi.
\newblock Unsupervised learning of landmarks by descriptor vector exchange.
\newblock In {\em Proceedings of the IEEE International Conference on Computer
  Vision}, pages 6361--6371, 2019.

\bibitem{thewlis2017unsupervised}
James Thewlis, Hakan Bilen, and Andrea Vedaldi.
\newblock Unsupervised learning of object frames by dense equivariant image
  labelling.
\newblock In {\em Advances in neural information processing systems}, pages
  844--855, 2017.

\bibitem{tompson2015efficient}
Jonathan Tompson, Ross Goroshin, Arjun Jain, Yann LeCun, and Christoph Bregler.
\newblock Efficient object localization using convolutional networks.
\newblock In {\em Proceedings of the IEEE conference on computer vision and
  pattern recognition}, pages 648--656, 2015.

\bibitem{heatmaps}
Jonathan~J Tompson, Arjun Jain, Yann LeCun, and Christoph Bregler.
\newblock Joint training of a convolutional network and a graphical model for
  human pose estimation.
\newblock In Z. Ghahramani, M. Welling, C. Cortes, N.~D. Lawrence, and K.~Q.
  Weinberger, editors, {\em Advances in Neural Information Processing Systems
  27}, pages 1799--1807. Curran Associates, Inc., 2014.

\bibitem{toshev2014deeppose}
Alexander Toshev and Christian Szegedy.
\newblock Deeppose: Human pose estimation via deep neural networks.
\newblock In {\em Proceedings of the IEEE conference on computer vision and
  pattern recognition}, pages 1653--1660, 2014.

\bibitem{tung2017adversarial}
Hsiao-Yu~Fish Tung, Adam~W Harley, William Seto, and Katerina Fragkiadaki.
\newblock Adversarial inverse graphics networks: Learning 2d-to-3d lifting and
  image-to-image translation from unpaired supervision.
\newblock In {\em 2017 IEEE International Conference on Computer Vision
  (ICCV)}, pages 4364--4372. IEEE, 2017.

\bibitem{wang2019adversarial}
Mengjiao Wang, Zhixin Shu, Shiyang Cheng, Yannis Panagakis, Dimitris Samaras,
  and Stefanos Zafeiriou.
\newblock An adversarial neuro-tensorial approach for learning disentangled
  representations.
\newblock {\em International Journal of Computer Vision}, 127(6-7):743--762,
  2019.

\bibitem{wei2016convolutional}
Shih-En Wei, Varun Ramakrishna, Takeo Kanade, and Yaser Sheikh.
\newblock Convolutional pose machines.
\newblock In {\em Proceedings of the IEEE conference on Computer Vision and
  Pattern Recognition}, pages 4724--4732, 2016.

\bibitem{wiles2018self}
Olivia Wiles, A Koepke, and Andrew Zisserman.
\newblock Self-supervised learning of a facial attribute embedding from video.
\newblock {\em arXiv preprint arXiv:1808.06882}, 2018.

\bibitem{pose_baselines}
Bin Xiao, Haiping Wu, and Yichen Wei.
\newblock Simple baselines for human pose estimation and tracking.
\newblock In {\em European Conference on Computer Vision (ECCV)}, 2018.

\bibitem{yang20183d}
Wei Yang, Wanli Ouyang, Xiaolong Wang, Jimmy Ren, Hongsheng Li, and Xiaogang
  Wang.
\newblock 3d human pose estimation in the wild by adversarial learning.
\newblock In {\em Proceedings of the IEEE Conference on Computer Vision and
  Pattern Recognition}, pages 5255--5264, 2018.

\bibitem{yang2011articulated}
Yi Yang and Deva Ramanan.
\newblock Articulated pose estimation with flexible mixtures-of-parts.
\newblock In {\em CVPR 2011}, pages 1385--1392. IEEE, 2011.

\bibitem{zhang}
Y. Zhang, Yijie Guo, Y. Jin, Yijun Luo, Zhiyuan He, and H. Lee.
\newblock Unsupervised discovery of object landmarks as structural
  representations.
\newblock {\em 2018 IEEE/CVF Conference on Computer Vision and Pattern
  Recognition}, pages 2694--2703, 2018.

\bibitem{puppet}
Silvia Zuffi and Michael~J. Black.
\newblock The stitched puppet: A graphical model of {3D} human shape and pose.
\newblock In {\em IEEE Conf. on Computer Vision and Pattern Recognition (CVPR
  2015)}, pages 3537--3546, June 2015.

\end{thebibliography}
}

\end{document}


\title{Unsupervised Human Pose Estimation through Transforming Shape Templates: Supplemental Material}

\author[1]{Luca Schmidtke}
\author[1]{Athanasios Vlontzos}
\author[1]{Simon Ellershaw}
\author[3]{Anna Lukens}
\author[2]{\\Tomoki Arichi}
\author[1]{and Bernhard Kainz}
\affil[1]{Imperial College London}
\affil[2]{King's College London}
\affil[3]{Evelina London Children's Hospital}

\maketitle
\section{Network Architectures}
In this section, we present the details regarding network architecture. Each block represents a layer. Abbreviations are as following: k=kernel size, s=stride, d = dilation, ch= number of channels, n=number of (fully connected) neurons.
\subsection{Our Approach}

\begin{figure}[htp]
\centering
\begin{minipage}{.5\textwidth}
  \centering
  \includegraphics[width=.7\linewidth]{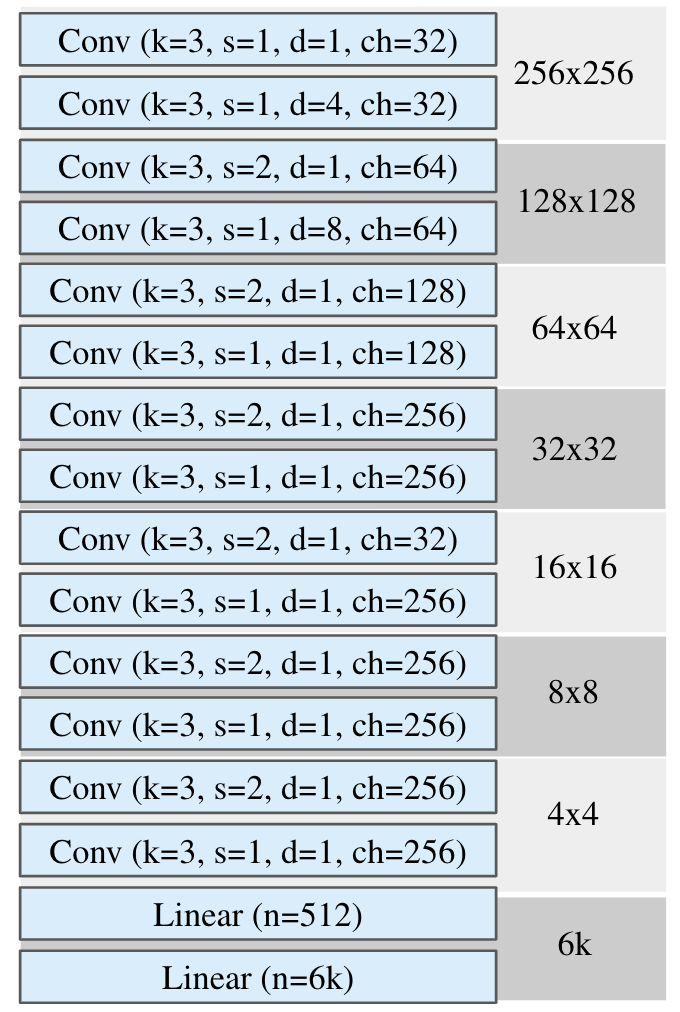}
  \captionof{figure}{Architecture of \textbf{pose extractor $\varphi$.} All layers\\ 
   are followed by batch normalisation and LeakyReLU \\
   except for the last.}
  \label{fig:test1}
\end{minipage}%
\begin{minipage}{.5\textwidth}
  \centering
  \includegraphics[width=.7\linewidth]{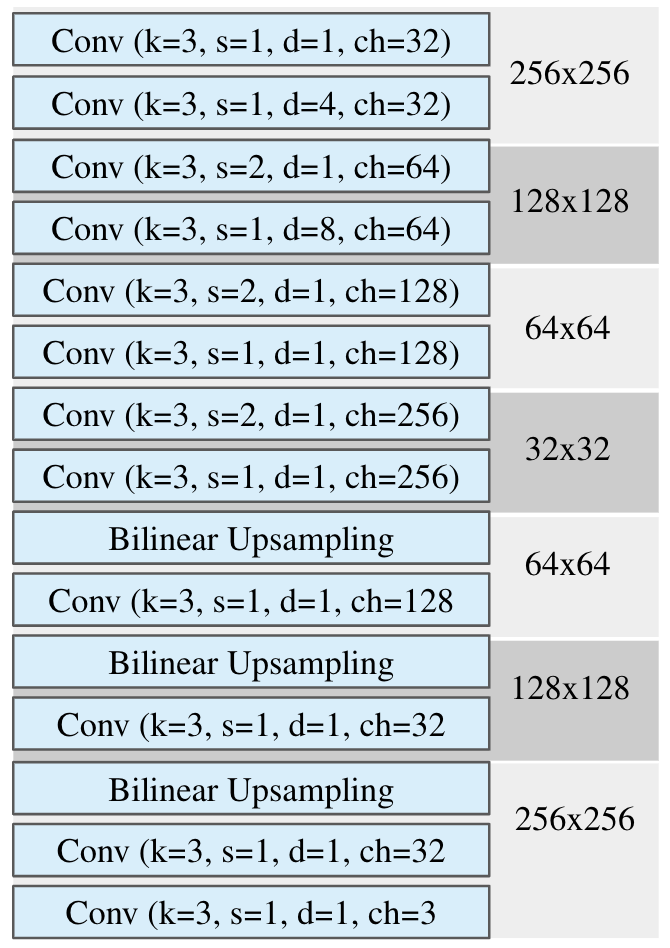}
  \captionof{figure}{Architecture of \textbf{image decoder $\phi$.} All layers \\
   are followed by batch normalisation and LeakyReLU \\
   except for the last.}
  \label{fig:test2}
\end{minipage}
\end{figure}
\clearpage
\subsection{Jakab et al. \cite{jakab_skeleton}}
For more details regarding the training procedure, refer to \cite{jakab_skeleton} and the accompanying supplemental material.

\begin{figure}[htp]
\centering
\begin{minipage}{.5\textwidth}
  \centering
  \includegraphics[width=.7\linewidth]{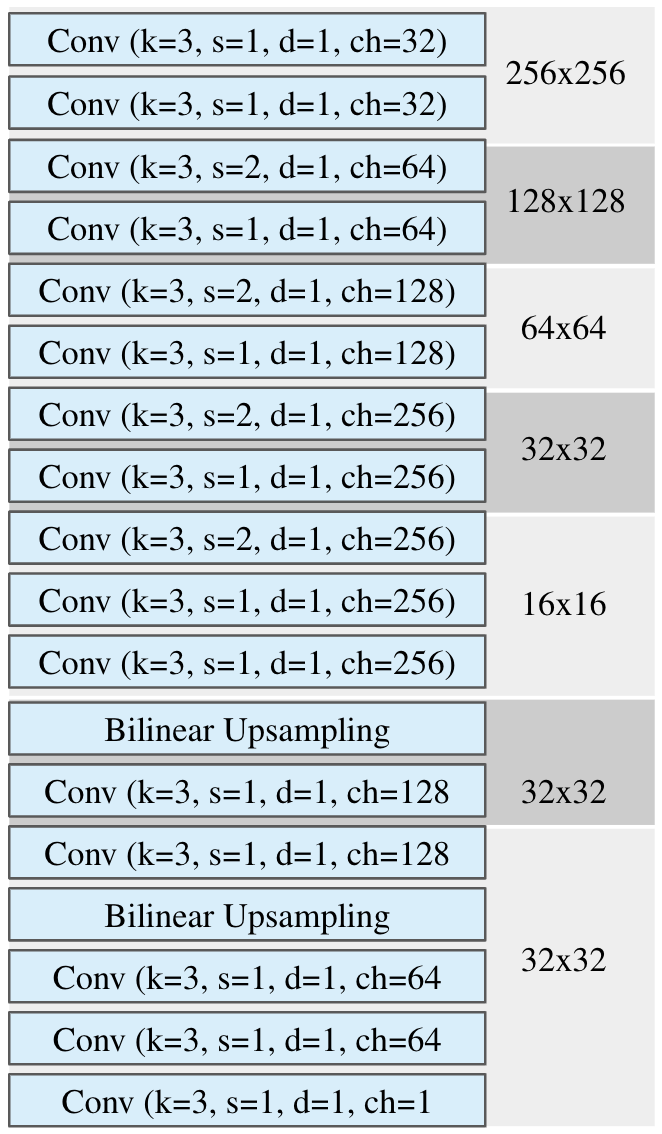}
  \captionof{figure}{Architecture of \textbf{skeleton encoder} All layers \\
  are followed by batch normalisation and LeakyReLU \\
  except for the last.}
  \label{fig:test1}
\end{minipage}%
\begin{minipage}{.5\textwidth}
  \centering
  \includegraphics[width=.9\linewidth]{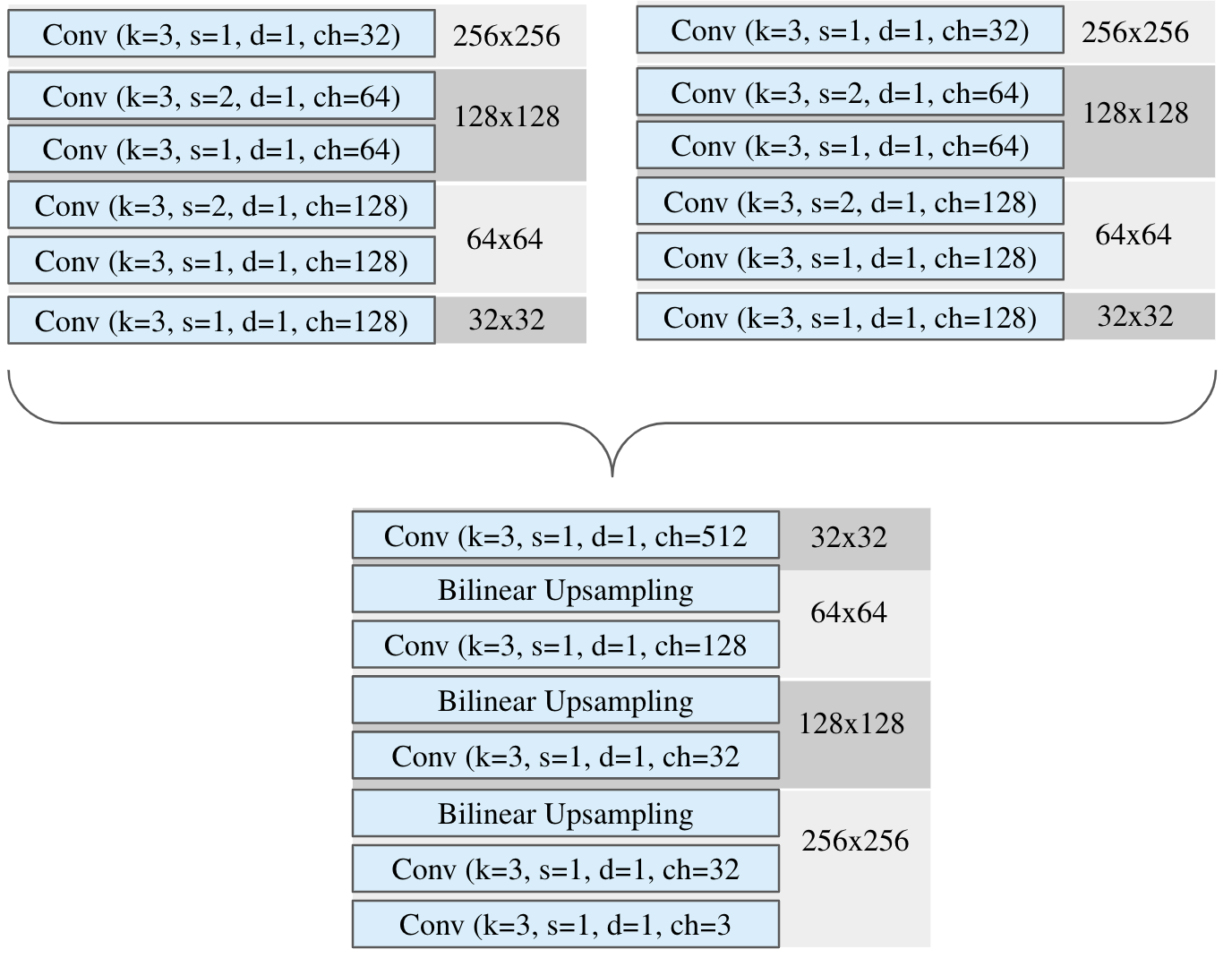}
  \captionof{figure}{Architecture of \textbf{image decoder.} All layers\\
  are followed by batch normalisation and LeakyReLU \\
  except for the last.}
  \label{fig:test2}
\end{minipage}
\end{figure}
\begin{figure}[htp]
\centering
\begin{minipage}{.5\textwidth}
  \centering
  \includegraphics[width=.7\linewidth]{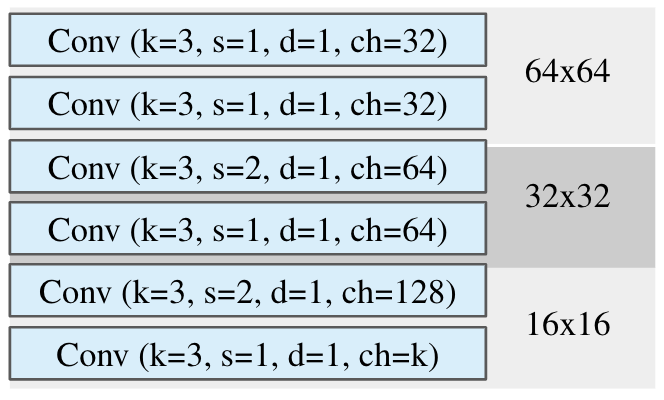}
  \captionof{figure}{Architecture of \textbf{skeleton regressor.} All layers \\
  are followed by batch normalisation and LeakyReLU \\
  except for the last.}
  \label{fig:test1}
\end{minipage}%
\begin{minipage}{.5\textwidth}
  \centering
  \includegraphics[width=.6\linewidth]{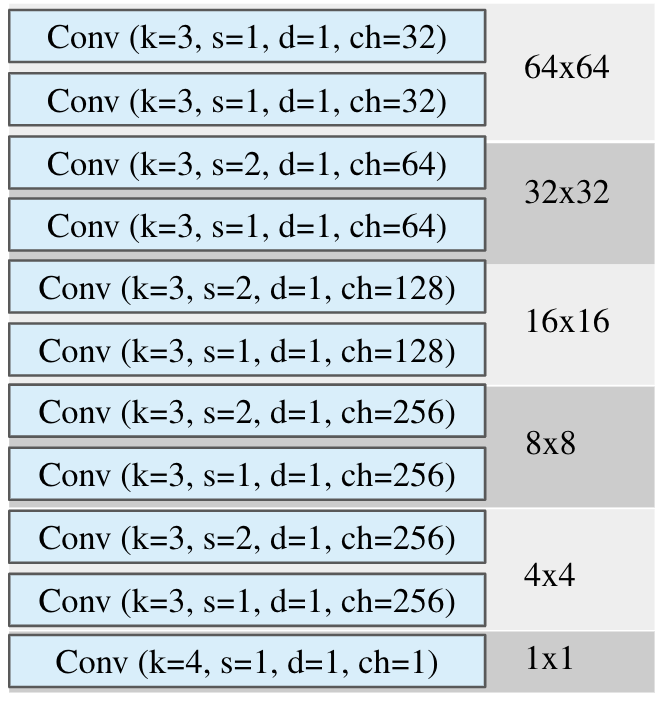}
  \captionof{figure}{Architecture of \textbf{the skeleton discriminator} \\
  All layers are followed by instance normalisation and \\ 
  LeakyReLU except for the last.}
  \label{fig:test2}
\end{minipage}
\end{figure}
\clearpage

\section{Results}
\subsection{Human3.6m}

\begin{figure}[h]
\begin{center}
  \includegraphics[width=0.85\linewidth]{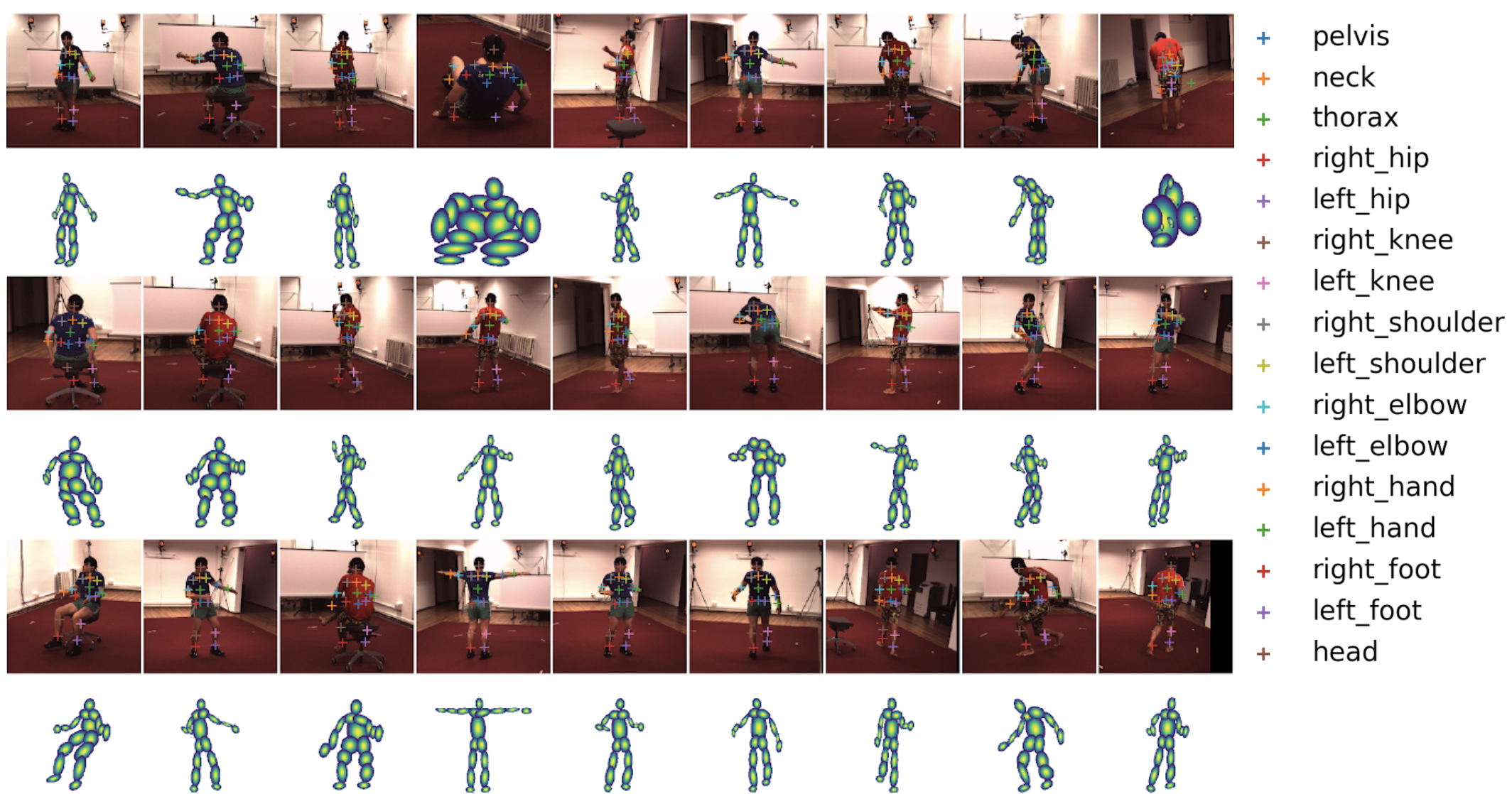}
\end{center}
\label{fig:results_infants}
\caption{Results for Human 3.6m. Markers denote detected landmarks.}
\end{figure}

\subsection{Infants}
\begin{figure}[h]
\begin{center}
  \includegraphics[width=0.85\linewidth]{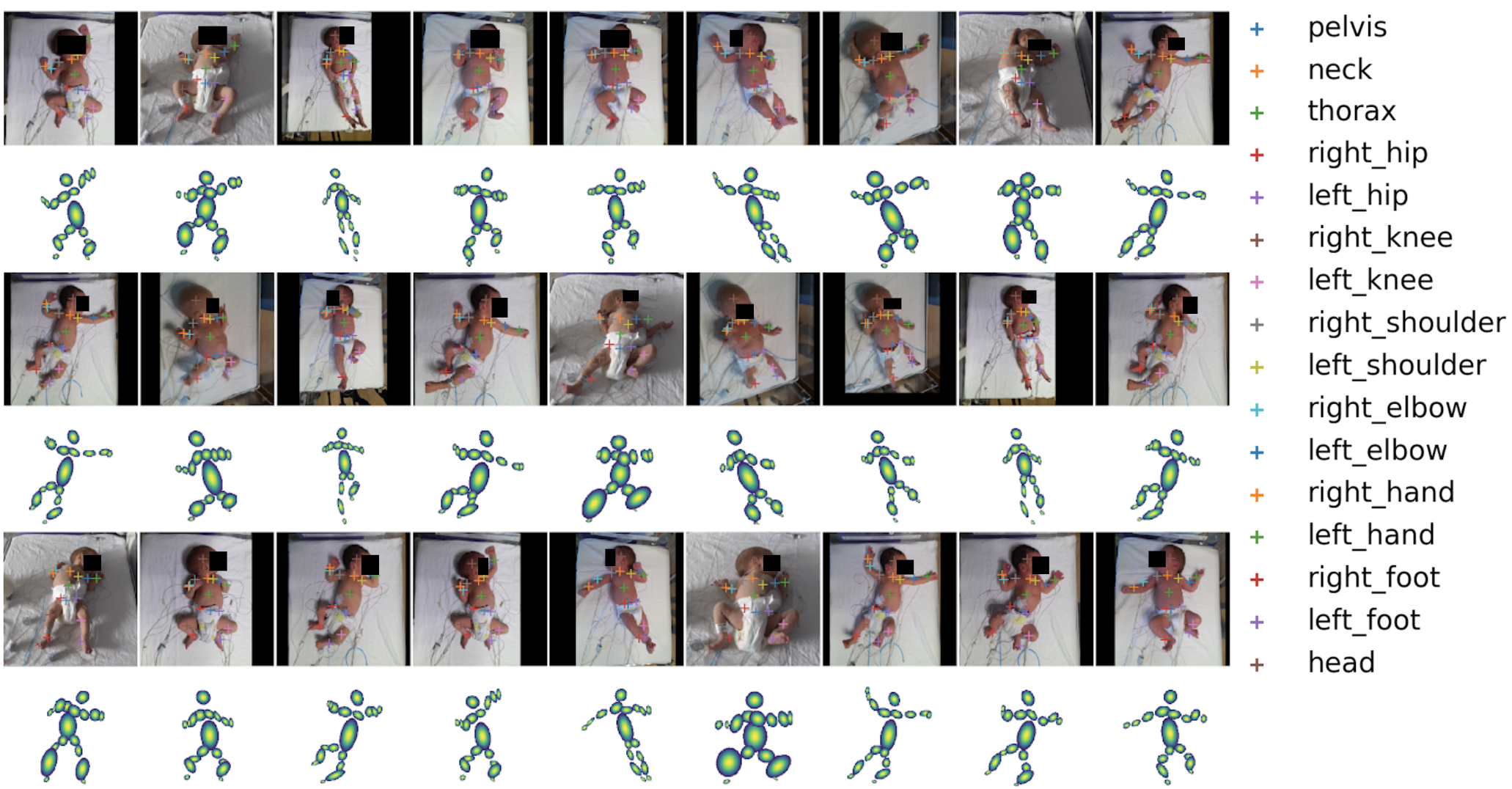}
\end{center}
\label{fig:results_infants}
\caption{Results for infants. Markers denote detected landmarks.}
\end{figure}

\section{Training and evaluation details}

\subsection{Human3.6m}
The model is trained with the Adam optimizer \cite{Adam}, a learning rate of 1e-4 and a batch size of 48. For evaluation, we choose 16 keypoints corresponding to following indices: 0, 13, 12, 1, 6, 2, 7, 25, 17, 26, 18, 27, 19, 3, 8, 14. While most of these keypoints conincide with our anchor points, we place two additional points on the thorax and the head in our default template.

\subsection{Infants}
For the infants, both models are trained the Aadam optimizer \cite{Adam}, a learning rate of 1e-4 and a batch size of 48. In order to train \cite{jakab_skeleton} with labels from the synthetic infants dataset, we project 3d coordinates into the image plane via a perspective camera transformation. Since the camera position in our clinical infant dataset is varying in terms of tilting angles, we augment the data via random rotations of the camera before projection in an effort to mimic these conditions.

\newpage
\bibliographystyle{ieee_fullname}
\bibliography{supp_bib}